\documentclass{article}
\usepackage{spconf,amsmath,graphicx}

\usepackage{times}
\usepackage{soul}
\usepackage{url}
\usepackage[hidelinks]{hyperref}
\usepackage[utf8]{inputenc}
\usepackage[small]{caption}
\usepackage{graphicx}
\usepackage{amsmath}
\usepackage{amsthm}
\usepackage{booktabs}
\usepackage{algorithm}
\usepackage{algorithmic}
\usepackage[switch]{lineno}

\usepackage{multirow,booktabs,subfigure}
\usepackage{amsmath}
\usepackage{subfigure}
\usepackage{makecell}
\usepackage{amssymb}

\title{Credible Teacher for Semi-Supervised Object Detection in Open Scene}

\name{Jingyu Zhuang \sthanks{These authors contributed equally. $^{\dagger}$Corresponding author.  This work was supported in part by the National Natural Science Foundation of China (NO.~62322608), in part by the Shenzhen Science and Technology Program (NO.~JCYJ20220530141211024), in part by the Open Project Program of the Key Laboratory of Artificial Intelligence for Perception and Understanding, Liaoning Province (AIPU, No.~20230003).
} \qquad Kuo Wang$^{\ast}$ \qquad Liang Lin \qquad Guanbin Li$^{\dagger}$}

\address{Sun Yat-sen University,  Guangzhou, China \\ School of Computer Science and Engineering, Research Institute of Sun Yat-sen University in Shenzhen}
\begin{document}
%\ninept
%
\maketitle
\begin{abstract}
Semi-Supervised Object Detection (SSOD) has achieved resounding success by leveraging unlabeled data to improve detection performance. 
However, in \textbf{Open Scene Semi-Supervised Object Detection} (\textbf{O-SSOD}), unlabeled data may contains unknown objects not observed in the labeled data, which will increase uncertainty in the model's predictions for known objects.
It is detrimental to the current methods that mainly rely on self-training, as more uncertainty leads to the lower localization and classification precision of pseudo labels.
To this end, we propose \textbf{Credible Teacher}, an end-to-end framework. 
Credible Teacher adopts an interactive teaching mechanism using flexible labels to prevent uncertain pseudo labels from misleading the model and gradually reduces its uncertainty through the guidance of other credible pseudo labels. 
Empirical results have demonstrated our method effectively restrains the adverse effect caused by O-SSOD and significantly outperforms existing counterparts.
\end{abstract}

\section{Introduction}
\label{sec:intro}

Recently, Semi-Supervised Object Detection (SSOD) methods have gained attention for leveraging massive unlabeled images to improve the performance of detectors when labeled data is limited. Existing SSOD methods assume labeled and unlabeled images share the same distribution and semantic categories to be detected, which is impractical for most real-world SSOD applications. 
Due to the complexity of real-world scenarios, unlabeled images often have different appearance distributions than carefully collected labeled data, and may include objects of unknown categories. 
Therefore, this paper focuses on a more realistic application scenario: Open Scene Semi-Supervised Object Detection (O-SSOD), where the differences in data distribution and object categories between labeled and unlabeled images may bring potential risks to existing methods based on pseudo labels.

As pointed out in~\cite{dhamija2020overlooked}, when an unlabeled image contains objects unseen in the labeled set, the model's localization accuracy and prediction confidence for known objects decrease. It is fatal for the current pseudo-labeling based SSOD methods, which mainly separate pseudo labels and background by setting a threshold.  
In O-SSOD, a higher threshold may guarantee the accuracy of pseudo labels, but ignores many known objects with lower confidence.
On the contrary, a lower threshold may lead to mislabeling more unknown objects as known.  
Moreover, inaccurate localization of object bounding boxes can also lead to incorrect assignment of pseudo labels. To sum up, unlabeled data collected from open scenes will introduce far more noisy pseudo labels in the training process, which will seriously damage the learning of the model.

In this paper, we propose an end-to-end framework \textbf{Credible Teacher} for O-SSOD, which is based on the teacher-student structure and contains a \emph{Teacher} and a \emph{Student}.
The \emph{Teacher} obtains credible pseudo labels on the weakly augmented image and labels the strongly augmented one to train the \emph{Student}. 
In order to reduce the fatal impact of noisy pseudo labels on model training in O-SSOD,  Credible Teacher adopts an interactive teaching mechanism with flexible labels. 
This mechanism enables the \emph{Teacher} to give a flexible label to each candidate box of the \emph{Student}, instead of directly assigning a one-hot pseudo label with the largest overlapping area. 
This can reduce additional noise caused by assigning pseudo labels with inaccurate localization. 
Flexible labels divide the class predictions of each candidate box into \emph{credible} and \emph{uncertain} through upper and lower thresholds. 
For the \emph{uncertain} predictions, we adopt a more conservative way by directly using them as soft labels to supervise the training of the corresponding object samples, which encourages the \emph{Student} to maintain consistent predictions with the \emph{Teacher} under different augmentations. 
In this way, Credible Teacher can prevent the \emph{Student} from being misguided by the incorrect predictions of the \emph{Teacher} and simultaneously enhance the robustness of the \emph{Student}.
For the \emph{credible} predictions, their values are set to 0 or 1, enhancing the \emph{Student}'s ability to recognize these objects in unlabeled images.

In addition, to alleviate the impact of cross-data distribution shift in O-SSOD, Credible Teacher adopts Data-specific Batch Normalization, as the labeled and unlabeled data under different distributions may have different characteristics that are not compatible within a single batch normalization layer.  
To validate the performance of our method in O-SSOD, we use MS-COCO and Objects365 datasets to construct a benchmark and Credible Teacher outperforms existing methods by a large margin in this open scene setting. 
Besides, in the traditional SSOD setting, Credible Teacher also performs well.

\begin{figure*}[t]
  \centering
  \includegraphics[width=0.75\textwidth]{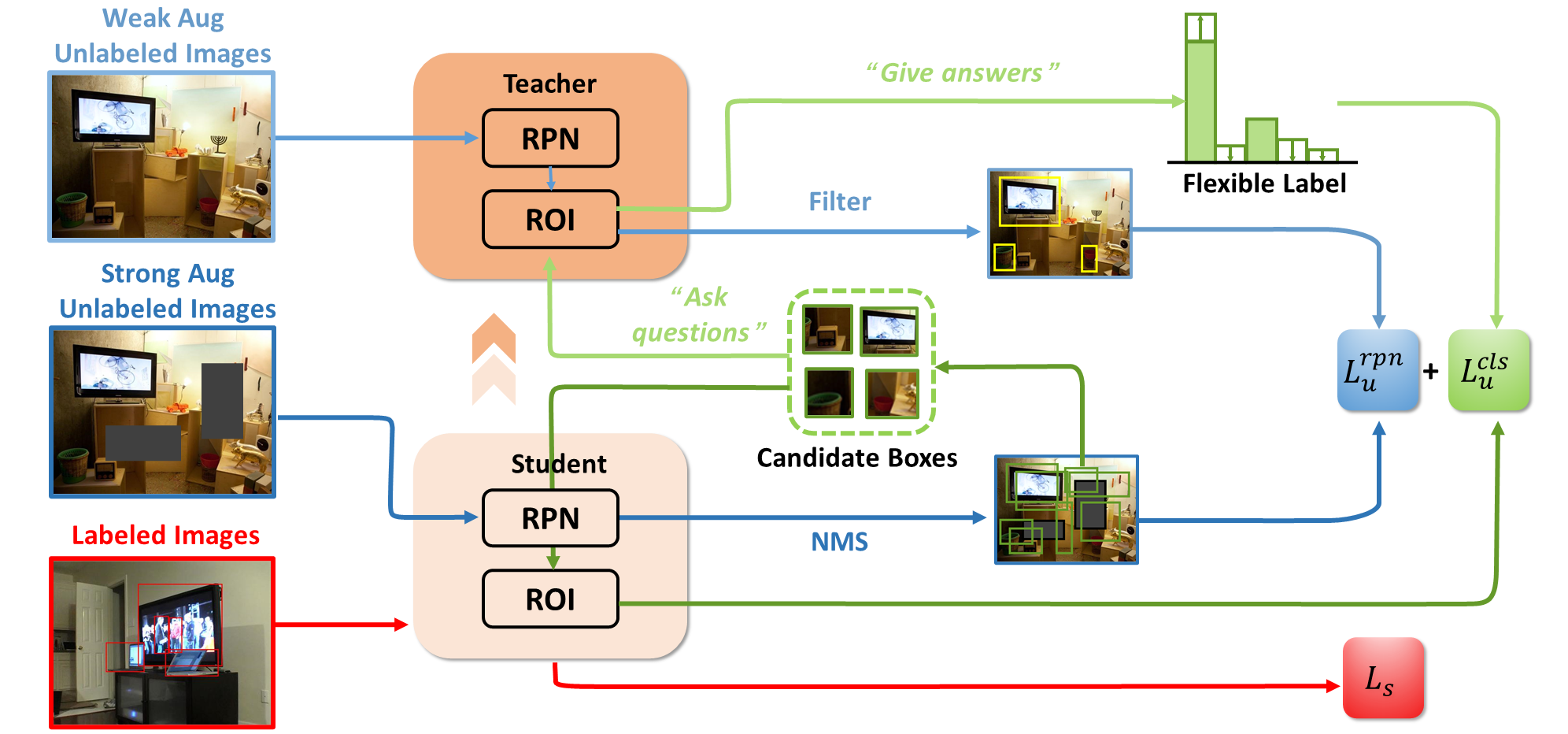}
  \caption{
The overall framework of Credible Teacher, which synchronously executes the supervised process and the unsupervised process to update the \emph{Student}, while the \emph{Teacher} is updated via the exponential moving average (EMA).
% In the supervised process, the \emph{Student} computes the supervised loss $L_{s}$ on the labeled images. 
% In the unsupervised process, the \emph{Teacher} first generates pseudo objects (yellow boxes) on weak augmented unlabeled images. Then, the \emph{Student} uses them to compute the rpn loss $L^{rpn}_{u}$ on strong augmented unlabeled images. 
% Next, we proceed to the interactive teaching mechanism. The \emph{Student} proposes candidate boxes and sends their locations to the \emph{Teacher}. The \emph{Teacher} generates flexible labels to compute the classification loss $L^{cls}_{u}$ of the \emph{Student}. 
% The \emph{Student} updates the \emph{Teacher} in the EMA manner.
}
  \label{fig:framework}
\end{figure*}

\section{Related works}
\label{sec:related}

SSOD has two mainstreams: consistency-based ones \cite{jeong2019consistency} and pseudo-label-based ones \cite{sohn2020simple,xu2021end, wang2023biased}. 
The former employ consistency losses by adding noise to images.
The pseudo-label-based methods have recently obtained impressive progress. STAC \cite{sohn2020simple} proposes to generate pseudo labels using weakly augmented images to train the model. 
Recently, many works \cite{xu2021end,liu2021unbiased,zheng2022dual,wang2022double, li2023gradient, zhang2023semi, nie2023adapting} adopt the teacher-student model to solve SSOD.
However, all these methods do not consider the O-SSOD setting~\cite{dhamija2020overlooked, zhuang2022discovering, zhuang2022open} widely exists in real-world applications, where the quality of pseudo labels  will be seriously degraded, thus deteriorating the performance of the model.

\section{Methodology}
\label{sec:method}

\textbf{Problem Setting.}
In O-SSOD, we have a labeled dataset $D_{s}=\left\{x^{s}_{i},y^{s}_{i}\right\}^{N_{s}}_{i=1}$ and an unlabeled dataset $D_{u}=\left\{x^{u}_{i}\right\}^{N_{u}}_{i=1}$. 
For each labeled image $x^{s}$, the corresponding label $y^{s}$ contains locations, sizes, and object categories of all bounding boxes. 
Let $\mathcal{C}_{s}$ denotes the category set of objects in labeled data, and $\mathcal{C}_{u}$ denotes that in unlabeled data. 
In O-SSOD, there exist objects of unknown classes ($\mathcal{C}_{s} / \mathcal{C}_{u} \neq \emptyset$) in unlabeled data. 
The goal of O-SSOD is to use both labeled and unlabeled data to learn a detection model.

\noindent
\textbf{The Overall Framework.}
As shown in the Fig.~\ref{fig:framework}, Credible Teacher is based on the Teacher-Student architecture~\cite{sohn2020simple} and adopts the two-stage detector Faster-RCNN~\cite{ren2015faster} including a region proposal subnetwork (RPN) and an ROI head. 
Credible Teacher simultaneously performs two training processes, i.e., the supervised process and the unsupervised process, to update the \emph{Student}, while the \emph{Teacher} is updated via the exponential moving average (EMA).

In the supervised process, the \emph{Student} takes a labeled sample ($x^{s}$, $y^{s}$) as input and computes the classification loss and the regression loss on the outputs of the RPN and ROI head. The supervised loss $\mathcal{L}_{s}$ is calculated as follows:
\begin{equation}
\mathcal{L}_{s} =  \sum_{i} \mathcal{L}_{cls}(x^{s}_{i},y^{s}_{i}) + \mathcal{L}_{reg}(x^{s}_{i},y^{s}_{i})  
\label{supervised loss}
\end{equation} 

In the unsupervised process, we employ strong and weak augmentation on an unlabeled image, obtaining $\ddot{x}^{u}_{i}$ and $\dot{x}^{u}_{i}$, respectively. 
For $\dot{x}^{u}_{i}$, the objects detected by the \emph{Teacher} with high confidence will be regarded as pseudo-labeled objects ${\hat{y}}^{u}_{i}$.
We use them to calculate the classification loss and regression loss on the output of the RPN in the \emph{Student}. 
For $\ddot{x}^{u}_{i}$, the \emph{Student} generates candidate boxes with RPN, and sends their locations to the \emph{Teacher} like asking questions.
In O-SSOD, the \emph{Teacher} may not give credible  ``answers''.
To prevent wrong ``answers'' from misguiding the \emph{Student}, the \emph{Teacher} uses flexible labels ${\tilde{y}}^{u}_{i}$ to calculate the classification loss on the output of the ROI head in the student branch. 
Unsupervised loss is thus defined as follows:
\begin{equation}
	\mathcal{L}_{u} = \sum_{i} \mathcal{L}^{rpn}_{cls}(\ddot{x}^{u}_{i},{\hat{y}}^{u}_{i}) + \mathcal{L}^{rpn}_{reg}(\ddot{x}^{u}_{i},{\hat{y}}^{u}_{i}) + \mathcal{L}^{roi}_{cls}(\ddot{x}^{u}_{i},{\tilde{y}}^{u}_{i}))
	\label{unsupervised loss}
\end{equation} 

We use a factor $\lambda$ to balance two training processes.

\noindent
\textbf{Interactive Teaching Mechanism with Flexible Labels.} 
The current pseudo-label-based methods mainly generate pseudo labels by filtering with a single threshold, and assign them to candidate boxes through IoU matching.
However, in O-SSOD, models trained on the labeled data tend to have greater localization and classification uncertainty for unlabeled data which brings two problems: 1) the pseudo labels may have large localization offsets from the ground truth. 2) the classification uncertainty makes it difficult to set a proper threshold to filter pseudo labels.

To solve these problems, Credible Teacher adopts an interactive teaching mechanism. After the \emph{Student} generates candidate boxes with RPN, unlike the mainstream methods of assigning the pseudo labels with the largest IoU to a candidate box, Credible Teacher sends the locations of the candidate boxes to the \emph{Teacher} to extract features from the weakly augmented images. Then the \emph{Teacher} directly generates flexible labels on the features to guide the \emph{Student}.
Assuming $n$ types of known objects, for each candidate box, the \emph{Teacher} divides its $n+1$ prediction results (the $(n+1)$-th prediction denotes the probability of background) into credible ones and uncertain ones, based on an upper threshold $\tau_{up}$ and a lower threshold $\tau_{low}$.
Specifically, prediction results with scores greater than the upper threshold or less than the lower threshold will be regarded as credible predictions and their values of flexible labels are set to 1 or 0, respectively. 
The remaining predictions are regarded as uncertain ones. In order to avoid the \emph{Teacher} wrongly labeling these candidate boxes and misleading the \emph{Student}, the value of the uncertain class predictions in the flexible labels will remain unchanged. For classes with uncertain predictions, flexible labels will not tell the \emph{Student} whether its prediction is correct but will encourage it to have the same predictions as the \emph{Teacher} under different augmentations. 
In this way, Credible Teacher avoids the pseudo labels assignment process that may introduce extra noise, and using the flexible labels to guide the candidate boxes can further reduce the impact of noise during training.

\noindent
\textbf{Data-specific Batch Normalization.}
As pointed out by~\cite{seo2020learning}, due to possible distribution shift between labeled and unlabeled data, batch normalization may not be optimal in O-SSOD. 
Motivated by~\cite{chang2019domain}, in Credible Teacher, we adopt Data-specific Batch Normalization (DBN) to mitigate the uncertainty of the model's prediction on unlabeled data caused by the cross-data distribution shift.

In DBN, we use two separate sets of means and variances for labeled and unlabeled data to compute their means and variances, respectively. Different from~\cite{chang2019domain} separating affine parameters, we share the affine parameters for labeled and unlabeled data. 
\cite{chang2019domain} firstly uses labeled data to train the model with the first set of affine parameter, and then extracts pseudo labels to finetune the model with the second one. 
However, in our end-to-end framework, the quality of flexible labels is inevitably poor at the beginning of training. If we only use the flexible labels to update the second set of affine parameters, the training process will be unstable, which will easily lead to the vanishing gradient problem. 
Therefore, we adopt shared affine parameters and train them using both labeled data and flexible labels from unlabeled data to ensure the training stability. Meanwhile, we still separate means and separate variances to solve the problem of distribution differences. 
DBN normalizes the activations from labeled and unlabeled data as follows:
\begin{equation}
    {\boldsymbol{\rm x}_{u}}' = \alpha \frac{\boldsymbol{\rm x}_{u}-\mu_{u}}{\sigma_{u}} + \beta \quad\quad {\boldsymbol{\rm x}_{l}}' = \alpha \frac{\boldsymbol{\rm x}_{l}-\mu_{l}}{\sigma_{l}} + \beta
	\label{DBN}
\end{equation}
where $\boldsymbol{\rm x}_{u}$ and $\boldsymbol{\rm x}_{l}$ denote the activations for labeled and unlabeled data, respectively; $(\mu_{u}, \sigma_{u})$ and $(\mu_{l}, \sigma_{l})$ are two independent sets of means and variances; $\alpha$ and $\beta$ are the shared affine parameters.

We replace all BN layers of the backbone with DBN. Since we do not know the distribution of test data, in testing, we use the mean of the labeled and unlabeled statistics of training data for inference, i.e., $\mu = \frac{\mu_{u}+\mu_{l}}{2}, \sigma_ = \frac{\sigma_{u}+\sigma_{l}}{2}$.

\section{Experiments}
\label{sec:experiments}

\textbf{Dataset.} We evaluate our method on two famous object detection datasets: the MS-COCO dataset \cite{lin2014microsoft} and the Objects365 dataset \cite{shao2019objects365}.
MS-COCO contains 80 object classes. It's subset \emph{train2017} is used as the label data. Objects365 comprises 365 categories, and we use Objects365 \emph{training} as the unlabeled data. 
After eliminating the same categories, Objects365 has 280 unique types of objects, \emph{i.e.}, unknown objects. Like \cite{sohn2020simple}, our experiments are divided into two protocols: 1) \textbf{partially labeled data protocol}, where we sample 1\%, 5\% and 10\% images of MS-COCO \emph{train2017} as the labeled data, and 10\% images of Objects365 \emph{training} as the unlabeled.  
2) \textbf{fully labeled data protocol}, where we use whole MS-COCO \emph{train2017} and Objects365 \emph{training} as the labeled and the unlabeled data, respectively. 

\noindent 
\textbf{Baselines and Evaluation Criteria.} 
We compare Credible Teacher with the current state-of-the-art SSOD methods: STAC~\cite{sohn2020simple}, Unbiased Teacher \cite{liu2021unbiased} and Soft teacher~\cite{xu2021end}.
We compute the mAP of the 80 known classes on both the MS-COCO val2017 set and the Objects365 validating set.

\noindent 
\textbf{Implementation setting.} 
We implement our method with MMdetection~\cite{chen2019mmdetection}.
In all experiments, we set the threshold $\sigma$ for filtering pseudo objects to 0.9 and set $\tau_{up}$ and $\tau_{low}$ to 0.8 and 0.05, respectively. The balance factor $\lambda$ is set to 2 and 4 in partially labeled and fully labeled data, respectively.
Following \cite{xu2021end}, we adopt the same weak and strong data augmentation schemes.
For partially labeled data, all methods are trained for 180K iterations with a batch size of 40, and the learning rate is 0.01. For fully labeled data, all methods are trained for 720K iterations with a batch size of 80 and the learning rate 0.02. We use mini-batch SGD for optimization and set the momentum as 0.9.

\begin{table*}[t]
  %\renewcommand\tabcolsep{5.0pt}
  % \small
  \caption{
  Results of MAP in the traditional SSOD and the O-SSOD setting. Supervised baseline is only trained with labeled COCO \emph{train2017}.
  }
  \label{tab:all}
  \begin{tabular}{l p{0.01cm}<{\centering} p{0.82cm}<{\centering} p{0.82cm}<{\centering} p{0.82cm}<{\centering} p{0.82cm}<{\centering} p{0.82cm}<{\centering} p{0.82cm}<{\centering} p{0.82cm}<{\centering} p{0.82cm}<{\centering} p{0.01cm} p{0.82cm}<{\centering} p{0.82cm}<{\centering} p{0.82cm}<{\centering} p{0.82cm}<{\centering}}
    \toprule
    \multirow{1}*{} & & \multicolumn{8}{c}{Results in O-SSOD setting } & & \multicolumn{4}{c}{Results in SSOD setting} \\
     \cmidrule{1-1} \cmidrule{3-10} \cmidrule{12-15} 
    \multirow{2}*{Method} & & \multicolumn{4}{c}{mAP on Objects365 \emph{validation} } & \multicolumn{4}{c}{mAP on COCO \emph{val2017} } & & \multicolumn{4}{c}{mAP on COCO \emph{val2017} } \\
    \cmidrule{3-10} \cmidrule{12-15} 
    & & 1\% & 5\% & 10\% & 100\% & 1\% & 5\% & 10\% & 100\% & & 1\% & 5\% & 10\%\\
    \cmidrule{1-1} \cmidrule{3-10} \cmidrule{12-15} 
    Supervised baseline & & 5.8 & 11.3 & 14.1 &23.1 & 10.7 & 20.3 & 24.5 &37.5 & & 10.0 & 20.9 & 26.9 \\
    STAC &  & 7.1 & 12.7  &  15.3 & 23.8 & 13.0  & 22.8  & 26.4  &38.7  & & 14.0  & 24.4  & 28.6 \\
    Unbiased teacher & & 9.6 & 14.5 & 15.9 & 24.2 & 16.5 & 25.1 & 28.6 & 40.7 & & 20.7 & 28.3 & 31.5\\
    Soft teacher & & 8.3 & 13.4 & 15.8  & 25.0 & 15.4 & 24.5  & 29.2 & 42.8 & & 20.5  & 30.7 & 34.0\\
    Ours & & \textbf{10.0} & \textbf{16.2} & \textbf{19.4} & \textbf{28.6} & \textbf{17.3} & \textbf{27.4} & \textbf{31.9} & \textbf{44.0} & & \textbf{23.5} & \textbf{31.7}  & \textbf{34.8}\\
     % \cmidrule{1-1} \cmidrule{3-10} \cmidrule{12-15} 
    \bottomrule
  \end{tabular}
\end{table*}

\noindent 
\textbf{Results under the traditional SSOD.}
First, we evaluate the performance of our method under the traditional SSOD setting, where 1\%, 5\% and 10\% of COCO \emph{train2017} are sampled as labeled data, and the remaining is unlabeled.
As shown in the right part of Table.\ref{tab:all}, our method achieves the best performance in all proportions.

\noindent 
\textbf{Results under the O-SSOD setting.}
In Table.\ref{tab:all}, we report the results of all methods on both partially labeled data and fully labeled data under the O-SSOD setting. Whether the test data is COCO \emph{val2017} or Objects365 \emph{validation}, our method achieves the best mAP in all proportions.  
When testing on Objects365~\emph{validation} set, due to the massive unknown objects in unlabeled data deteriorating the quality of pseudo labels, most current SSOD methods only achieve a limited performance improvement. In contrast, our method outperforms the supervised baseline by~\textbf{5.5} mAP and outperforms the second best method Soft teacher by~\textbf{3.6} mAP on fully labeled data.
It can be concluded that our proposed method can alleviate the introduction of pseudo-label noise and mine more helpful information for model training from unlabeled data w.r.t the more realistic O-SSOD setting, thus obtaining a larger performance boost. 
Besides, unlike STAC and Soft teacher, which perform poorly with 1\% labeled data, our method has stable performance gains on different proportions of labeled data. This shows that our method is effective for different proportions of labeled data, which is important in practical applications.

\begin{table}[t]
  \centering
  \small
  \caption{Ablation studies of our Credible Teacher.
  }
  \label{tab:Ablation}
  % \small
  \begin{tabular}{l p{1.cm}<{\centering} p{1.cm}<{\centering} p{1.cm}<{\centering}}
    \toprule
      \quad  Flexible labels  & IMT & DBN & mAP \\
    \midrule    
    (1)  & & & 14.8 \\
    (2)  \quad\quad \checkmark  & & &  16.1 \\
    (3)    & \checkmark & &  15.6 \\
    (4)     & & \checkmark & 15.9 \\
    (5) \quad\quad  \checkmark & \checkmark & &  18.1 \\
    (6) \quad\quad  \checkmark & \checkmark & \checkmark &  \textbf{19.4} \\
    \bottomrule
  \end{tabular}
\end{table}

\begin{figure}
  \centering
  \includegraphics[width=1.0 \columnwidth]{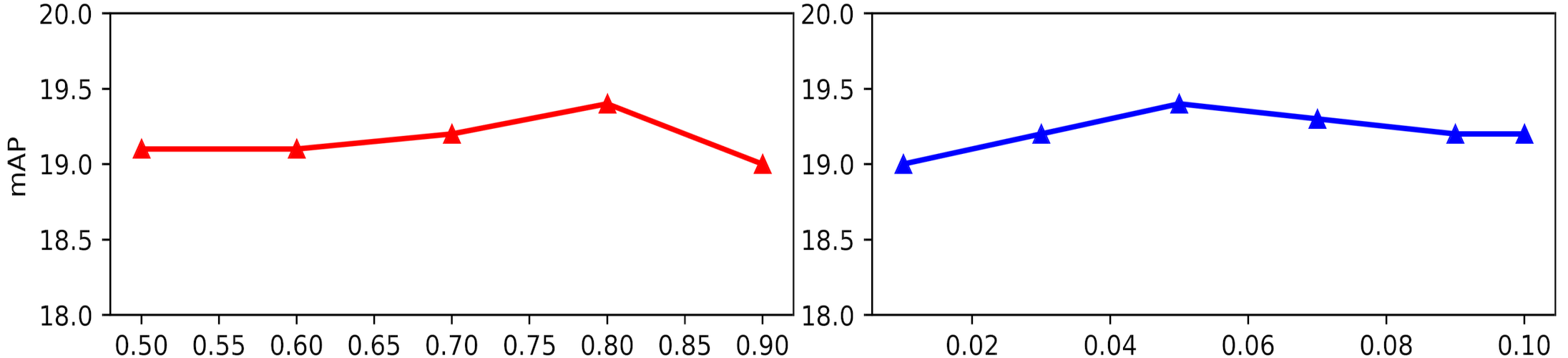}
  \caption{Sensitivity of the upper threshold $\tau_{up}$ and the lower threshold $\tau_{low}$. Left: we fix $\tau_{low}=0.05$ and vary the $\tau_{up}$ from 0.5 to 0.9. Right: we fix $\tau_{up}=0.8$ and vary the $\tau_{low}$ from 0.01 to 0.1.}
  \label{fig:Sensitivity}
\end{figure}

\noindent 
\textbf{Ablation study.}
The contributions of different components of Credible Teacher are listed in Table.\ref{tab:Ablation}. IMT indicates the interactive teaching mechanism. In (1), (3) and (4), we use soft pseudo labels, which are common used in many works. In (1), (2), (3), and (5), we all use common BN.
From (1) to (4), we find that each component of Credible Teacher will improve the model's performance, but the improvement is limited.
However, comparing (1) with (5), when we combine the interactive teaching mechanism with flexible labels, Credible Teacher gains a huge performance improvement. It proves that, while avoiding introducing more noisy pseudo labels, the interactive teaching mechanism with flexible labels can give enough guidance to the \emph{Student} through credible class prediction results. In other words, it enhances the model's recognition of objects and gradually reduces the uncertainty of the model's prediction on unlabeled data. 
Pairs (1)-(4) and (5)-(6) indicate that DBN can alleviate the problem caused by cross-data distribution shift and bring stable performance improvements in the O-SSOD setting.

\noindent 
\textbf{Sensitivity analysis of thresholds}.
We investigate the sensitivity of the two thresholds by counting the performance of Credible Teacher with different values of  $\tau_{up}$ and $\tau_{low}$. The results are shown in Fig.\ref{fig:Sensitivity}, which shows that the performance of Credible Teacher slightly floats in $ \pm$0.2. Flexible labels are proved to be insensitive to hyperparameters $\tau_{up}$ and $\tau_{low}$.

\section{Conclusions}
\label{sec:conclusions}

In this paper, we focus on O-SSOD, a more realistic semi-supervised object detection scenario which boosts the mislabeling risk for the current pseudo-labels-based models and leads to a negative effect. 
To tackle this challenging problem, we propose Credible Teacher, which uses the interactive teaching mechanism with flexible labels to prevent the noise pseudo labels from further misguiding the model and gradually improve the quality of pseudo labels. Besides, Credible Teacher employs data-specific batch normalization to alleviate the impact of the cross-data distribution shift. 
We conduct experiments to verify the threat of O-SSOD on the existing methods and demonstrate the superior performance of our proposed Credible Teacher.

\bibliographystyle{IEEEbib}
\bibliography{Template}

\end{document}